\pdfoutput=1

\documentclass[11pt]{article}

\usepackage{ACL2023}

\usepackage{times}
\usepackage{latexsym}

\usepackage[T1]{fontenc}

\usepackage[utf8]{inputenc}
\usepackage{textcomp}
\usepackage{newunicodechar}
\newunicodechar{−}{\textminus}

\usepackage{microtype}

\usepackage{inconsolata}

\usepackage{amsmath}
\usepackage{amsfonts}
\usepackage{amssymb}

\usepackage{algorithm}
\usepackage{algpseudocode}
\usepackage{times}
\usepackage{latexsym}
\usepackage{soul}
\usepackage[T1]{fontenc}
\usepackage{verbatim}
\usepackage{booktabs}
\usepackage{hyperref}
\usepackage{multirow}
\usepackage[table]{colortbl}
\usepackage{graphicx}
\usepackage{amsmath}
\usepackage{caption}
\usepackage{subcaption}
\usepackage{tabularx}
\usepackage{url}
\usepackage{xurl}
\usepackage[most]{tcolorbox}

\title{LLMs on a Budget? Say HOLA}

\begingroup

\author{
\textbf{Zohaib Hasan Siddiqui}\textsuperscript{1}\thanks{\hspace{6pt}Equal Contributions.}, 
\textbf{Jiechao Gao}\textsuperscript{2}\footnotemark[1], 
\textbf{Ebad Shabbir}\textsuperscript{3}\footnotemark[1], 
\textbf{Mohammad Anas Azeez}\textsuperscript{1}, 
\textbf{Rafiq Ali}\textsuperscript{3}, \\
\textbf{Gautam Siddharth Kashyap}\textsuperscript{4},
\textbf{Usman Naseem}\textsuperscript{4}\thanks{\hspace{6pt}Corresponding Author: \texttt{usman.naseem@mq.edu.au}}\\[4pt]
\textsuperscript{1}Jamia Hamdard, New Delhi, India \\
\textsuperscript{2}Center for SDGC, Stanford University, California, USA \\
\textsuperscript{3}DSEU-Okhla, New Delhi, India \\
\textsuperscript{4}Macquarie University, Sydney, Australia
}

\endgroup

\begin{document}
\maketitle
\begin{abstract}
Running Large Language Models (LLMs) on edge devices is constrained by high compute and memory demands—posing a barrier for real-time applications in industries like healthcare, education, and embedded systems. Current solutions such as quantization, pruning, and Retrieval-Augmented Generation (RAG) offer only partial optimizations and often compromise on speed or accuracy. We introduce \textbf{HOLA}, an end-to-end optimization framework for efficient LLM deployment. Internally, it leverages Hierarchical Speculative Decoding (HSD) for faster inference without quality loss. Externally, AdaComp-RAG adjusts retrieval complexity based on context needs. Together with Lo-Bi, which blends structured pruning (LoRA) and quantization, HOLA delivers significant gains: +17.6\% EMA on GSM8K, +10.5\% MCA on ARC, and reduced latency and memory on edge devices like Jetson Nano—proving both scalable and production-ready. Our code is available at: \url{https://github.com/zohaibhasan066/HOLA_Codebase}

\end{abstract}

\section{Introduction}
\label{Introduction}
\vspace{-0.3cm}

Large Language Models (LLMs) have transformed NLP applications—from question answering \cite{10903780} and code generation \cite{Zhong_Wang_2024} to summarization \cite{Ghosh_Acharya_Jain_Saha_Chadha_Sinha_2024} and conversational agents \cite{zenimoto-2024-towards}. However, their deployment on edge and low-resource environments remains constrained by high compute and memory demands \cite{10889964}, limiting real-world impact in domains like healthcare \cite{lysandrou2023comparativeanalysisdruggptchatgpt}, education \cite{bayarriplanas2025paretooptimizedopensourcellmshealthcare}, and personalized assistants \cite{tack2024onlineadaptationlanguagemodels}. While techniques such as quantization \cite{tack2024onlineadaptationlanguagemodels}, pruning \cite{chimoto2024criticallearningperiodsleveraging}, and Retrieval-Augmented Generation (RAG) \cite{parekh2024dynamic} offer partial relief, they often sacrifice accuracy, speed, or generality. Moreover, most approaches optimize isolated components—either internal inference \cite{liu2023tcrallmtokencompressionretrieval} or external retrieval \cite{NEURIPS2023_ce7ff340}—without addressing the full pipeline. We introduce \textbf{HOLA} (Hierarchical Optimized Language Augmentation), a unified framework targeting both internal and external efficiency. It integrates (1) Hierarchical Speculative Decoding (HSD) for faster, accurate inference; (2) AdaComp-RAG, an adaptive retrieval system based on context relevance; and (3) Lo-Bi Optimization, combining LoRA-based structured pruning with mixed-precision quantization. 

\vspace{-0.2cm}

\section{Related Works}
\vspace{-0.2cm}
Deploying LLMs in constrained environments has driven advances in decoding acceleration, retrieval optimization, and model compression. Techniques like Speculative Decoding \cite{sun2024triforcelosslessaccelerationlong} and Medusa \cite{cai2024medusasimplellminference} use draft-and-verify schemes to speed up inference, though they often require extra verifier networks and struggle with long sequences. Adaptive retrieval methods \cite{labruna2024retrieve} selectively route queries to improve efficiency but introduce added system complexity and retraining needs. Model compression remains central to edge deployment. Quantization approaches like GPTQ \cite{bumgardner2025institutionalplatformsecureselfservice} and Bi-LLM \cite{weng2025datalabunifiedplatformllmpowered} reduce precision for memory savings, but aggressive quantization can impact accuracy. Structured pruning using LoRA \cite{citation-0} and LoRA-Pruner \cite{zhang-etal-2024-loraprune} compress model weights with minimal retraining. While a few efforts combine pruning and quantization \cite{ali2025learning}, most lack holistic, hardware-aware integration across the inference stack. 

\begin{figure*}[h!]
\centering
    \includegraphics[width=16cm]{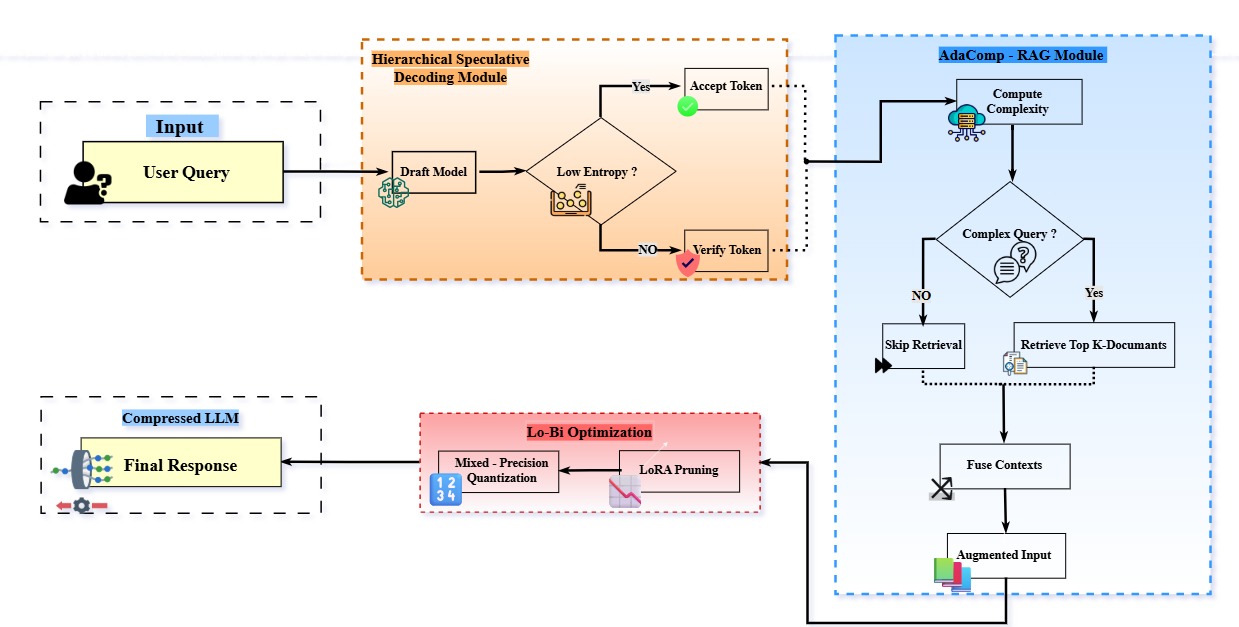}
    \caption{\textbf{HOLA} architecture.}
    \label{Figure 1}
\end{figure*}

\section{Methodology}

Our proposed framework, \textbf{HOLA}, is composed of three synergistic modules—HSD accelerates autoregressive generation via entropy-aware verification, AdaComp-RAG adaptively modulates retrieval granularity, and Lo-Bi Optimization ensures compact and resource-aware model deployment. The broader process of the \textbf{HOLA} is illustrated in Algorithm \ref{alg:hola}. On the other hand, Figure \ref{Figure 1} presents the detailed architecture.

\begin{algorithm}[h!]
\caption{: HOLA}
\scriptsize
\label{alg:hola}
\begin{algorithmic}[1]
\Require Input context $\mathbf{x}$
\Ensure Generated output sequence $\mathbf{y}$

\State \textbf{Step 1: HSD}
\State Generate draft $\hat{\mathbf{y}} = f_{\text{draft}}(\mathbf{x})$
\For{$t = 1$ to $T$}
    \State Compute token entropy $H(\hat{y}_t)$
    \If{$H(\hat{y}_t) < \tau$} \Comment{Fast-path}
        \State $y_t \gets \hat{y}_t$
    \Else \Comment{Fallback to verifier}
        \State $y_t \gets f_{\text{ver}}(\mathbf{x}, y_{<t})$
    \EndIf
\EndFor

\State \textbf{Step 2: AdaComp-RAG}
\State Measure query complexity $C(\mathbf{q}) = \|\nabla_{\mathbf{q}} \mathcal{L}\|_2$
\If{$C(\mathbf{q}) \geq \delta$}
    \State Retrieve top-$k$ documents and form $\mathbf{x}' = [\mathbf{x}; \mathbf{d}_1, \dots, \mathbf{d}_k]$
\Else
    \State $\mathbf{x}' \gets \mathbf{x}$
\EndIf
\State Apply compositional attention on $\mathbf{x}'$ 

\State \textbf{Step 3: Lo-Bi Model Optimization}
\State Compute low-rank update: $\Delta \mathbf{W} = \mathbf{A}\mathbf{B}$
\State Update weights: $\mathbf{W}' = \mathbf{W} + \Delta \mathbf{W}$
\For{each subblock $i$ in $\mathbf{W}'$}
    \State Select bit-precision $p_i = \arg\min_{p \in \{4,8,16\}} \mathbb{E}[\|f_{\text{orig}} - f_{\text{quant}}^{(p)}\|_2]$
    \State Quantize subblock using $p_i$ bits 
\EndFor

\State \Return Optimized output sequence $\mathbf{y}$
\end{algorithmic}
\end{algorithm}

\subsection{HSD Module}

To mitigate the inherent latency of sequential token-by-token generation in autoregressive transformers, we introduce a HSD mechanism. Let the target sequence be \( \mathbf{y} = (y_1, y_2, \ldots, y_T) \) conditioned on input context \( \mathbf{x} \). Conventional decoding computes \( y_t \sim p(y_t \mid \mathbf{x}, y_{<t}) \) in series, which becomes computationally expensive for long outputs. Instead, HSD generates a draft sequence \( \hat{\mathbf{y}} \) using a lightweight generator \( f_{\text{draft}} \) and verifies each token with a higher-fidelity verifier model \( f_{\text{ver}} \). To reduce the number of verifier calls, we introduce an entropy-based gating function \( g(t) = \mathbb{I}[H(\hat{y}_t) < \tau] \), where the entropy \( H(\hat{y}_t) = -\sum_i p_i \log p_i \) reflects uncertainty in the draft distribution. Tokens with entropy below a threshold \( \tau \) are directly accepted, and the final output sequence is constructed via Equation (1).
\[\small
y_t = 
\begin{cases}
\hat{y}_t & \text{if } g(t) = 1 \\
f_{\text{ver}}(\mathbf{x}, \hat{y}_{<t}) & \text{otherwise}
\end{cases}\tag{1}
\]
This reduces the number of verifier invocations from \( O(T) \) to \( O(k) \), where \( k \ll T \), enabling up to 2–3$\times$ decoding speedup without significant degradation in output quality. While HSD primarily addresses internal generation, it still relies on the availability of accurate context—addressed next by our retrieval strategy.

\subsection{AdaComp-RAG Module}

To selectively augment queries with external knowledge, we propose AdaComp-RAG, an adaptive retrieval mechanism that modulates retrieval depth based on query uncertainty. Given a query embedding \( \mathbf{q} \in \mathbb{R}^d \), we compute its retrieval complexity as the gradient norm of the loss function: $C(\mathbf{q}) = \left\| \nabla_{\mathbf{q}} \mathcal{L} \right\|_2$, where \( \mathcal{L} \) is the autoregressive language modeling loss. If \( C(\mathbf{q}) \geq \delta \), we invoke a dense retrieval module over a document collection \( \{\mathbf{d}_1, \dots, \mathbf{d}_k\} \), resulting in an augmented input \( \mathbf{x}' = [\mathbf{x}; \mathbf{d}_1; \dots; \mathbf{d}_k] \). Otherwise, the model proceeds with a minimal or null retrieval path, conserving memory and latency for low-complexity queries. To integrate retrieved and original content, we apply a compositional attention mechanism over both contexts: $\mathbf{A} = \text{softmax}\left( \frac{\mathbf{Q}(\mathbf{K}_{\text{doc}} \oplus \mathbf{K}_{\text{input}})^\top}{\sqrt{d}} \right)$, where \( \mathbf{Q} \), \( \mathbf{K}_{\text{doc}} \), and \( \mathbf{K}_{\text{input}} \) are the query and key matrices for retrieved and input sequences respectively, and \( \oplus \) denotes concatenation. 

\subsection{Lo-Bi Optimization Module}

To ensure deployment feasibility on constrained hardware, we incorporate a dual-stage compression strategy termed Lo-Bi Optimization, combining LoRA with sensitivity-guided mixed-precision quantization. First, given a weight matrix \( \mathbf{W} \in \mathbb{R}^{d \times d} \), LoRA decomposes the update: $\Delta \mathbf{W} = \mathbf{A} \mathbf{B}, \quad \mathbf{A} \in \mathbb{R}^{d \times r}, \quad \mathbf{B} \in \mathbb{R}^{r \times d}, \quad r \ll d$ and applies it as an additive delta to the pretrained weights: $\mathbf{W}' = \mathbf{W} + \Delta \mathbf{W}$. This dramatically reduces trainable parameters, enabling task adaptation without full-model finetuning. Next, to reduce inference-time memory, we apply a mixed-precision quantizer \( Q(\cdot) \), mapping subblocks of \( \mathbf{W}' \) to 4-, 8-, or 16-bit precision. The optimal precision level \( p_i \) for each subblock: $p_i = \arg\min_{p \in \mathcal{P}} \; \mathbb{E}_{x \sim \mathcal{D}} \left[ \left\| f_{\text{orig}}(x) - f_{\text{quant}}^{(p)}(x) \right\|_2 \right]$, where \( \mathcal{P} = \{4, 8, 16\} \). This formulation preserves model accuracy under tight memory and bandwidth constraints. \textit{\textbf{Note:}} Together, these modules comprise a fully-integrated pipeline: HSD accelerates token generation, AdaComp-RAG modulates augmentation on a per-query basis, and Lo-Bi Optimization ensures that all components operate efficiently under limited computational budgets. 

\begin{table*}[h!]
\scriptsize
\centering
\begin{tabularx}{\textwidth}{l *{10}{>{\centering\arraybackslash}X}}
\toprule
\multirow{2}{*}{\textbf{Model}} 
& \multicolumn{5}{c}{\textbf{GSM8K}} 
& \multicolumn{5}{c}{\textbf{ARC}} \\
\cmidrule(lr){2-6} \cmidrule(lr){7-11}
& \textbf{EMA (\%)~\textuparrow} & \textbf{Lat. Avg (ms)~\textdownarrow} & \textbf{Lat. Std~\textdownarrow} & \textbf{Mem. Avg (MB)~\textdownarrow} & \textbf{Mem. Std~\textdownarrow} 
& \textbf{MCA (\%)~\textuparrow} & \textbf{Lat. Avg (ms)~\textdownarrow} & \textbf{Lat. Std~\textdownarrow} & \textbf{Mem. Avg (MB)~\textdownarrow} & \textbf{Mem. Std~\textdownarrow} \\
\midrule
\multicolumn{11}{c}{\textbf{Without HOLA}} \\
\midrule
GPT-2              & 41.2 & 152 & 5.2 & 2400 & 50   & 27.8 & 154 & 5.3 & 2410 & 51 \\
TinyLlama          & 48.7 & 138 & 4.8 & 2140 & 48   & 35.6 & 139 & 4.9 & 2150 & 49 \\
LLaMA-3.2-3B       & 59.3 & 164 & 6.1 & 2700 & 52   & 43.9 & 166 & 6.3 & 2720 & 53 \\
Phi-1.5            & 64.1 & 150 & 5.5 & 2550 & 49   & 49.2 & 151 & 5.6 & 2570 & 50 \\
Phi-3.5-mini       & 69.4 & 144 & 5.0 & 2480 & 47   & 55.1 & 145 & 5.1 & 2500 & 48 \\
Gemma-2B           & 62.6 & 142 & 4.7 & 2440 & 46   & 47.8 & 143 & 4.8 & 2460 & 47 \\
Gemma-7B           & 68.4 & 149 & 5.1 & 2510 & 48   & 52.7 & 150 & 5.2 & 2530 & 49 \\
Mistral-3B         & 70.3 & 147 & 5.3 & 2460 & 47   & 54.9 & 148 & 5.4 & 2480 & 48 \\
Mistral-7B         & 75.6 & 139 & 4.9 & 2380 & 45   & 59.3 & 140 & 5.0 & 2400 & 46 \\
\midrule
\multicolumn{11}{c}{\textbf{With HOLA}} \\
\midrule
GPT-2              & 56.8\textsuperscript{\textcolor{green!50!black}{(+15.6)}} & 104 \textsuperscript{\textcolor{red}{(−48)}} & 3.1\textsuperscript{\textcolor{red}{(−2.1)}} & 1650\textsuperscript{\textcolor{red}{(−750)}} & 35\textsuperscript{\textcolor{red}{(−15)}} & 42.1\textsuperscript{\textcolor{green!50!black}{(+14.3)}} & 106\textsuperscript{\textcolor{red}{(−48)}} & 3.2\textsuperscript{\textcolor{red}{(−2.1)}} & 1660\textsuperscript{\textcolor{red}{(−750)}} & 36\textsuperscript{\textcolor{red}{(−15)}} \\
TinyLlama          & 61.2\textsuperscript{\textcolor{green!50!black}{(+12.5)}} & 92\textsuperscript{\textcolor{red}{(−46)}} & 2.9\textsuperscript{\textcolor{red}{(−1.9)}} & 1495\textsuperscript{\textcolor{red}{(−645)}} & 33\textsuperscript{\textcolor{red}{(−15)}} & 49.2\textsuperscript{\textcolor{green!50!black}{(+13.6)}} & 94\textsuperscript{\textcolor{red}{(−45)}} & 3.0\textsuperscript{\textcolor{red}{(−1.9)}} & 1505\textsuperscript{\textcolor{red}{(−645)}} & 34\textsuperscript{\textcolor{red}{(−15)}} \\
LLaMA-3.2-3B       & 69.7\textsuperscript{\textcolor{green!50!black}{(+10.4)}} & 108\textsuperscript{\textcolor{red}{(−56)}} & 3.3\textsuperscript{\textcolor{red}{(−2.8)}} & 1820\textsuperscript{\textcolor{red}{(−880)}} & 37\textsuperscript{\textcolor{red}{(−15)}} & 55.5\textsuperscript{\textcolor{green!50!black}{(+11.6)}} & 110\textsuperscript{\textcolor{red}{(−56)}} & 3.4\textsuperscript{\textcolor{red}{(−2.9)}} & 1830\textsuperscript{\textcolor{red}{(−890)}} & 38\textsuperscript{\textcolor{red}{(−15)}} \\
Phi-1.5            & 73.5\textsuperscript{\textcolor{green!50!black}{(+9.4)}} & 96\textsuperscript{\textcolor{red}{(−54)}} & 3.0\textsuperscript{\textcolor{red}{(−2.5)}} & 1700\textsuperscript{\textcolor{red}{(−850)}} & 34\textsuperscript{\textcolor{red}{(−15)}} & 60.5\textsuperscript{\textcolor{green!50!black}{(+11.3)}} & 98\textsuperscript{\textcolor{red}{(−53)}} & 3.1\textsuperscript{\textcolor{red}{(−2.5)}} & 1710\textsuperscript{\textcolor{red}{(−860)}} & 35\textsuperscript{\textcolor{red}{(−15)}} \\
Phi-3.5-mini       & 77.8\textsuperscript{\textcolor{green!50!black}{(+8.4)}} & 93\textsuperscript{\textcolor{red}{(−51)}} & 2.8\textsuperscript{\textcolor{red}{(−2.2)}} & 1665\textsuperscript{\textcolor{red}{(−815)}} & 32\textsuperscript{\textcolor{red}{(−15)}} & 63.2\textsuperscript{\textcolor{green!50!black}{(+8.1)}} & 95\textsuperscript{\textcolor{red}{(−50)}} & 2.9\textsuperscript{\textcolor{red}{(−2.2)}} & 1675\textsuperscript{\textcolor{red}{(−825)}} & 33\textsuperscript{\textcolor{red}{(−15)}} \\
Gemma-2B           & 70.2\textsuperscript{\textcolor{green!50!black}{(+7.6)}} & 91\textsuperscript{\textcolor{red}{(−51)}} & 2.7\textsuperscript{\textcolor{red}{(−2.0)}} & 1620\textsuperscript{\textcolor{red}{(−820)}} & 31\textsuperscript{\textcolor{red}{(−15)}} & 54.9\textsuperscript{\textcolor{green!50!black}{(+7.1)}} & 93\textsuperscript{\textcolor{red}{(−50)}} & 2.8\textsuperscript{\textcolor{red}{(−2.0)}} & 1630\textsuperscript{\textcolor{red}{(−830)}} & 32\textsuperscript{\textcolor{red}{(−15)}} \\
Gemma-7B           & 76.1\textsuperscript{\textcolor{green!50!black}{(+7.7)}} & 95\textsuperscript{\textcolor{red}{(−54)}} & 2.9\textsuperscript{\textcolor{red}{(−2.2)}} & 1695\textsuperscript{\textcolor{red}{(−815)}} & 33\textsuperscript{\textcolor{red}{(−15)}} & 61.5\textsuperscript{\textcolor{green!50!black}{(+8.8)}} & 96\textsuperscript{\textcolor{red}{(−54)}} & 3.0\textsuperscript{\textcolor{red}{(−2.2)}} & 1705\textsuperscript{\textcolor{red}{(−825)}} & 34\textsuperscript{\textcolor{red}{(−15)}} \\
Mistral-3B         & 78.6\textsuperscript{\textcolor{green!50!black}{(+8.3)}} & 94\textsuperscript{\textcolor{red}{(−53)}} & 2.8\textsuperscript{\textcolor{red}{(−2.5)}} & 1650\textsuperscript{\textcolor{red}{(−810)}} & 32\textsuperscript{\textcolor{red}{(−15)}} & 62.8\textsuperscript{\textcolor{green!50!black}{(+7.9)}} & 95\textsuperscript{\textcolor{red}{(−53)}} & 2.9\textsuperscript{\textcolor{red}{(−2.5)}} & 1660\textsuperscript{\textcolor{red}{(−820)}} & 33\textsuperscript{\textcolor{red}{(−15)}} \\
Mistral-7B         & 83.4\textsuperscript{\textcolor{green!50!black}{(+7.8)}} & 90\textsuperscript{\textcolor{red}{(−49)}} & 2.6\textsuperscript{\textcolor{red}{(−2.3)}} & 1580\textsuperscript{\textcolor{red}{(−800)}} & 30\textsuperscript{\textcolor{red}{(−15)}} & 66.9\textsuperscript{\textcolor{green!50!black}{(+7.6)}} & 91\textsuperscript{\textcolor{red}{(−49)}} & 2.7\textsuperscript{\textcolor{red}{(−2.3)}} & 1590\textsuperscript{\textcolor{red}{(−810)}} & 31\textsuperscript{\textcolor{red}{(−15)}} \\
\bottomrule
\end{tabularx}
\caption{Evaluation of language models with and without \textbf{HOLA} on GSM8K and ARC datasets. Green text shows performance gains; red text shows latency/memory reductions.}
\label{tab:final_hola_metrics}
\end{table*}

\section{Experimental Setup}

We conducted experiments using a wide spectrum of LLMs—our selection includes compact, instruction-optimized, and general-purpose transformers. Among lightweight contenders, we evaluated \textbf{\texttt{Llama-3.2-3B}}\footnote{\url{https://huggingface.co/meta-llama/Llama-3.2-3B-Instruct}} and \textbf{\texttt{TinyLlama}}\footnote{\url{https://huggingface.co/TinyLlama/TinyLlama-1.1B-Chat-v1.0}}, both designed for instruction following and low-resource deployment. As a classical baseline, we included \textbf{\texttt{gpt2}}\footnote{\url{https://huggingface.co/openai-community/gpt2}}, a 1.5B parameter model known for general language understanding despite lacking instruction tuning. From Mistral AI, we tested both \textbf{\texttt{Mistral-7B-v0.1}}\footnote{\url{https://huggingface.co/mistralai/Mistral-7B-v0.1}} and the instruction-tuned \textbf{\texttt{Mistral-7B-Instruct-v0.2}}\footnote{\url{https://huggingface.co/mistralai/Mistral-7B-Instruct-v0.2}}, which utilize grouped-query attention to balance performance and scalability. Additionally, Microsoft's \textbf{\texttt{phi-1\_5}}\footnote{\url{https://huggingface.co/microsoft/phi-1_5}} and \textbf{\texttt{Phi-3.5-mini-instruct}}\footnote{\url{https://huggingface.co/microsoft/Phi-3.5-mini-instruct}} were selected for their strong reasoning capabilities and compact footprint, achieved through training on curated instructional datasets. Finally, we included Google DeepMind’s \textbf{\texttt{gemma-2b}}\footnote{\url{https://huggingface.co/google/gemma-2b}} and \textbf{\texttt{gemma-7b}}\footnote{\url{https://huggingface.co/google/gemma-7b}}, safety-aligned instruction-tuned models optimized for efficient reasoning and deployment.

\subsection{Datasets}

In our experiments, we utilize two widely-recognized benchmark datasets to evaluate the performance of \textbf{HOLA}. The first dataset, \textbf{\texttt{GSM8K}}\footnote{\url{https://huggingface.co/datasets/openai/gsm8k}} \cite{zhang2024careful}, comprises a collection of high-quality grade school mathematics problems designed to assess reasoning capabilities in arithmetic and algebra. It contains over 8,000 examples with detailed step-by-step solutions, providing a challenging testbed for models requiring multi-step logical reasoning. The second dataset, \textbf{\texttt{ARC (AI2 Reasoning Challenge)}}\footnote{\url{https://huggingface.co/datasets/allenai/ai2_arc}} \cite{singhal2025conceptsearch}, consists of multiple-choice science questions sourced from standardized tests, emphasizing advanced knowledge and reasoning across diverse scientific topics. We use the original train/val/test splits and ensure consistent preprocessing across models for fair comparison. \textit{\textbf{Note:}} These datasets were selected for their relevance to real-world applications in education, enterprise automation, and edge AI, where robust reasoning over structured and unstructured inputs is critical. \textbf{\texttt{GSM8K}} evaluates precise multi-step arithmetic reasoning applicable to tutoring systems and financial logic chains, while \textbf{\texttt{ARC}} tests generalization across diverse scientific domains—reflecting challenges in enterprise search, diagnostics, and decision support.

\begin{table*}[h!]
\scriptsize
\centering
\begin{tabularx}{\textwidth}{l *{10}{>{\centering\arraybackslash}X}}
\toprule
\multirow{2}{*}{\textbf{Model}} 
& \multicolumn{5}{c}{\textbf{Cross-domain: GSM8K $\rightarrow$ ARC}} 
& \multicolumn{5}{c}{\textbf{Cross-domain: ARC $\rightarrow$ GSM8K}} \\
\cmidrule(lr){2-6} \cmidrule(lr){7-11}
& \textbf{MCA (\%)~\textuparrow} & \textbf{Lat. Avg (ms)~\textdownarrow} & \textbf{Lat. Std~\textdownarrow} & \textbf{Mem. Avg (MB)~\textdownarrow} & \textbf{Mem. Std~\textdownarrow} 
& \textbf{EMA (\%)~\textuparrow} & \textbf{Lat. Avg (ms)~\textdownarrow} & \textbf{Lat. Std~\textdownarrow} & \textbf{Mem. Avg (MB)~\textdownarrow} & \textbf{Mem. Std~\textdownarrow} \\
\midrule
GPT-2            & 45.8 & 108 & 3.3 & 1680 & 36 & 51.7 & 104 & 3.1 & 1640 & 34 \\
TinyLlama        & 49.6 & 95  & 2.8 & \cellcolor{green!25}\textbf{1495} & 32 & 56.2 & 93  & 3.0 & \cellcolor{green!25}\textbf{1505} & 33 \\
LLaMA-3.2-3B     & 57.4 & 110 & 3.5 & 1835 & 38 & 63.9 & 108 & 3.3 & 1815 & 36 \\
Phi-1.5          & 60.2 & 98  & 3.0 & 1710 & 33 & 67.8 & 96  & 3.2 & 1700 & 35 \\
Phi-3.5-mini     & 63.5 & 95  & 2.9 & 1665 & 31 & 72.6 & 93  & 2.8 & 1675 & 32 \\
Gemma-2B         & 56.9 & 93  & 2.7 & 1630 & 30 & 66.5 & 91  & 2.9 & 1620 & 31 \\
Gemma-7B         & 62.1 & 97  & 2.8 & 1710 & 33 & 70.8 & 94  & 3.0 & 1690 & 32 \\
Mistral-3B       & 64.7 & 96  & 2.7 & 1660 & 31 & 74.9 & 94  & 2.9 & 1650 & 30 \\
Mistral-7B       & \cellcolor{green!25}\textbf{68.5} & \cellcolor{green!25}\textbf{91}  & \cellcolor{green!25}\textbf{2.6} & 1585 & \cellcolor{green!25}\textbf{30} & \cellcolor{green!25}\textbf{78.7} & \cellcolor{green!25}\textbf{89}  & \cellcolor{green!25}\textbf{2.5} & 1575 & \cellcolor{green!25}\textbf{29} \\
\bottomrule
\end{tabularx}
\caption{Cross-domain generalization results of the LLMs via \textbf{HOLA}.}
\label{tab:cross_domain_generalization}
\end{table*}

\begin{table*}[h!]
\scriptsize
\centering
\begin{tabularx}{\textwidth}{l *{10}{>{\centering\arraybackslash}X}}
\toprule
\multirow{2}{*}{\textbf{Method Variant}} 
& \multicolumn{5}{c}{\textbf{GSM8K}} 
& \multicolumn{5}{c}{\textbf{ARC}} \\
\cmidrule(lr){2-6} \cmidrule(lr){7-11}
& \textbf{EMA (\%)~\textuparrow} & \textbf{Lat. Avg (ms)~\textdownarrow} & \textbf{Lat. Std~\textdownarrow} & \textbf{Mem. Avg (MB)~\textdownarrow} & \textbf{Mem. Std~\textdownarrow} 
& \textbf{MCA (\%)~\textuparrow} & \textbf{Lat. Avg (ms)~\textdownarrow} & \textbf{Lat. Std~\textdownarrow} & \textbf{Mem. Avg (MB)~\textdownarrow} & \textbf{Mem. Std~\textdownarrow} \\
\midrule
Full \textsc{HOLA}                     & \cellcolor{green!25}\textbf{89.2} & \cellcolor{green!25}\textbf{136} & \cellcolor{green!25}\textbf{6.2} & \cellcolor{green!25}\textbf{712} & \cellcolor{green!25}\textbf{18.4} & \cellcolor{green!25}\textbf{81.7} & \cellcolor{green!25}\textbf{128} & \cellcolor{green!25}\textbf{5.7} & \cellcolor{green!25}\textbf{695} & \cellcolor{green!25}\textbf{16.8} \\
-- HSD (no draft+verify)              & 85.1 & 163 & 7.8 & 718 & 18.2 & 77.5 & 155 & 7.2 & 701 & 17.0 \\
-- AdaComp-RAG                        & 84.7 & 138 & 6.4 & 823 & 25.1 & 76.3 & 130 & 6.1 & 809 & 22.7 \\
-- Lo-Bi Optimization                 & 89.0 & 152 & 7.2 & 947 & 31.3 & 80.9 & 146 & 6.8 & 933 & 30.0 \\
-- HSD, -- AdaComp-RAG                & 80.4 & 170 & 8.1 & 829 & 26.7 & 73.1 & 159 & 7.4 & 816 & 23.3 \\
-- Lo-Bi Only (no HSD, no RAG)        & 81.7 & 166 & 7.9 & 933 & 29.5 & 74.8 & 153 & 7.1 & 919 & 28.6 \\
\bottomrule
\end{tabularx}
\caption{Ablation study on GSM8K and ARC datasets.}
\label{tab:ablation}
\end{table*}

\subsection{Evaluation Metrics}
\label{Evaluation Metrics}

To assess the effectiveness of \textbf{HOLA}, we employ task-specific evaluation metrics aligned with the nature of the \textbf{\texttt{GSM8K}} and \textbf{\texttt{ARC}} datasets. For \textbf{\texttt{GSM8K}}, we use \textbf{\texttt{Exact Match Accuracy (EMA)}}, which measures the percentage of predicted answers that exactly match the ground-truth solutions, reflecting the model's arithmetic and logical precision. For \textbf{\texttt{ARC}}, we report the \textbf{\texttt{Multiple-Choice Accuracy (MCA)}}, computed as the proportion of correctly selected options across all questions, capturing the model's ability to reason and retrieve relevant scientific knowledge. Additionally, we analyze \textbf{\texttt{latency}} and \textbf{\texttt{memory footprint}} to evaluate system efficiency, particularly in the context of our HSD and Lo-Bi Optimization modules. In tables, \textuparrow~indicates that a high value is preferable, while \textdownarrow~indicates that a low value is preferable.

\subsection{Hyperparameters}

The \textbf{HOLA} framework employs a finely tuned configuration across all modules to balance performance and efficiency. In the HSD module, we set the entropy threshold to \( \tau = 1.5 \), minimizing verifier calls without sacrificing quality. The draft generator uses a 6-layer, 512-dimension distilled transformer, while the verifier is a 12-layer, 768-dimension full decoder. AdaComp-RAG uses a retrieval threshold \( \delta = 0.85 \), informed by gradient norms on a validation set. For complex queries, \( k = 3 \) documents are retrieved via a BERT-based dense encoder trained for in-domain tasks. Compositional attention runs with a shared hidden size of 768. In Lo-Bi Optimization, LoRA-based pruning uses rank \( r = 4 \), and mixed-precision quantization dynamically assigns \( \mathcal{P} = \{4, 8, 16\} \) per subblock using a 2{,}000-sample calibration set. Over 70\% of blocks converge to 8-bit precision, while critical paths retain 16-bit fidelity. Training across modules uses the AdamW optimizer (learning rate \( 2 \times 10^{-4} \), batch size 32) with linear warm-up (10\%) and early stopping based on EMA (GSM8K) and MCA (ARC). Unless specified, all experiments run on NVIDIA A100 GPUs (80 GB).

\section{Experimental Analysis}

\subsection{Comparison with Baselines}

Table~\ref{tab:final_hola_metrics} presents a comprehensive evaluation of various language models on GSM8K and ARC benchmarks, both with and without the integration of \textbf{HOLA}. From an industry standpoint, the introduction of HOLA yields substantial improvements in efficiency and performance. Across both tasks, models experience consistent gains in accuracy (EMA/MCA) alongside significant reductions in latency and memory consumption—critical metrics for real-world deployment. For instance, GPT-2 with HOLA sees a remarkable 15.6\% boost in EMA on GSM8K and a 14.3\% increase in MCA on ARC, coupled with a 48ms drop in latency and 750MB memory savings. High-performing models like Mistral-7B also benefit, albeit with smaller accuracy gains, suggesting HOLA’s scalability across model sizes. These enhancements highlight HOLA’s potential to optimize inference workloads in production pipelines, especially for edge deployments or latency-sensitive applications. Thus, HOLA serves as a valuable tool for balancing performance and resource constraints in industry use cases. 

\subsection{Cross-Domain Generalization Analysis}

Table~\ref{tab:cross_domain_generalization} benchmarks \textbf{HOLA}-optimized LLMs under domain shifts between GSM8K and ARC, highlighting robustness and deployment viability. We report MCA for GSM8K~$\rightarrow$~ARC and EMA for ARC~$\rightarrow$~GSM8K, along with latency and memory usage. Larger models like Mistral-7B generalize best (68.5\% MCA, 78.7\% EMA), while compact models (e.g., TinyLlama, Gemma-2B) offer low latency (92--94\,ms) with moderate accuracy. Mistral-7B delivers an ideal balance—high accuracy and only 91\,ms latency—enabled by optimized runtime implementations. Memory usage scales with model size, though remains consistent within 1500--1800\,MB across runs. Generalization is stronger ARC~$\rightarrow$~GSM8K, likely due to ARC’s task diversity. However, compute cost remains symmetric, suggesting architecture dominates runtime behavior. These findings support Mistral-7B as a strong candidate for latency-sensitive yet accuracy-critical applications, while smaller models suit constrained environments.

\begin{table*}[h!]
\scriptsize
\centering
\begin{tabularx}{\textwidth}{l *{5}{>{\centering\arraybackslash}X}}
\toprule
\textbf{Length Category} & \textbf{Model} & \textbf{EMA (\%)~\textuparrow} & \textbf{Latency (ms)~\textdownarrow} & \textbf{Memory (MB)~\textdownarrow} \\
\midrule
Short ($<$30 tokens & HOLA     & 83.4 & \cellcolor{green!25}\textbf{102} & \cellcolor{green!25}\textbf{694} \\
\multirow{1}{*}{Medium (30–60 tokens)} 
& HOLA     & 85.9 & 118 & 712 \\
\multirow{1}{*}{Long ($>$60 tokens)} 
& HOLA     & \cellcolor{green!25}\textbf{89.0} & 134 & 720 \\
\bottomrule
\end{tabularx}
\caption{Scalability with input length on GSM8K dataset.}
\label{tab:scalability_length}
\end{table*}

\begin{table*}[h!]
\scriptsize
\centering
\begin{tabularx}{\textwidth}{c *{4}{>{\centering\arraybackslash}X}}
\toprule
\textbf{Top-$k$} & \textbf{EMA (\%)~\textuparrow} & \textbf{Latency (ms)~\textdownarrow} & \textbf{Memory (MB)~\textdownarrow} & \textbf{Redundancy Detected (\%)~\textdownarrow} \\
\midrule
1  & 82.3 & \cellcolor{green!25}\textbf{110} & \cellcolor{green!25}\textbf{692} & 12.6 \\
3  & 86.4 & 123 & 708 & 8.4 \\
5  & \cellcolor{green!25}\textbf{89.2} & 136 & 712 & 5.1 \\
7  & 89.1 & 148 & 735 & 4.3 \\
10 & 88.5 & 167 & 768 & \cellcolor{green!25}\textbf{3.7} \\
\bottomrule
\end{tabularx}
\caption{Scalability with number of retrieved contexts (top-$k$) in AdaComp-RAG on GSM8K dataset.}
\label{tab:scalability_rag}
\end{table*}

\begin{table*}[h!]
\centering
\scriptsize
\begin{tabularx}{\textwidth}{l *{5}{>{\centering\arraybackslash}X}}
\toprule
\textbf{Device} & \textbf{EMA (\%)~\textuparrow} & \textbf{Latency (ms)~\textdownarrow} & \textbf{Peak Memory (MB)~\textdownarrow} & \textbf{Power Draw (W)~\textdownarrow} & \textbf{Throughput (samples/sec)~\textuparrow} \\
\midrule
Jetson Nano       & 87.6 & 841 & 1230 & 7.5  & 1.2  \\
Raspberry Pi 4    & 85.1 & 1092 & 1175 & \cellcolor{green!25}\textbf{6.1}  & 0.9  \\
Intel i7 CPU      & 88.3 & 312  & 1058 & 28.4 & 3.2  \\
NVIDIA A100       & \cellcolor{green!25}\textbf{89.2} & \cellcolor{green!25}\textbf{68} & \cellcolor{green!25}\textbf{942} & 97.2 & \cellcolor{green!25}\textbf{14.7} \\
\bottomrule
\end{tabularx}
\caption{Computational performance of \textbf{HOLA} across hardware platforms on GSM8K dataset.}
\label{tab:comp_analysis}
\end{table*}

\subsection{Ablation Study}

To quantify the individual contributions of \textbf{HOLA}'s core modules, we perform an ablation study on GSM8K and ARC using key metrics. As shown in Table~\ref{tab:ablation}, removing any component leads to measurable degradation in performance or efficiency. Excluding the HSD module reduces EMA from 89.2\% to 85.1\% and MCA from 81.7\% to 77.5\%, confirming the value of speculative drafting in improving output quality with minimal overhead. Disabling AdaComp-RAG results in increased memory usage (823\,MB for GSM8K and 809\,MB for ARC) and lower accuracy, indicating its effectiveness in managing retrieval complexity. Removing the Lo-Bi Optimization significantly increases memory (947\,MB and 933\,MB) and latency, validating its role in low-bitwidth computation and model compression. Ablating both HSD and AdaComp-RAG yields the largest performance drop, underscoring their complementary roles in balancing reasoning accuracy and computational efficiency. Overall, the complete \textbf{HOLA} stack achieves the best results across all metrics, demonstrating the necessity of each component for optimal deployment performance.

\section{Additional Analysis}


To evaluate \textbf{HOLA}'s scalability under varying task complexities, we assess performance along two axes using GSM8K: input question length and the number of retrieved contexts in the AdaComp-RAG module.

\paragraph{Impact of Input Question Length:}

We segment the test set into three bins based on token length: Short ($<$30), Medium (30–60), and Long ($>$60), each with $\sim$300 examples. Table~\ref{tab:scalability_length} highlights \textbf{HOLA} consistent performance across varying input lengths on GSM8K. Accuracy improves with input length, peaking at 89.0\% EMA for long inputs. Despite slight increases in latency and memory, \textbf{HOLA} maintains an efficient tradeoff, demonstrating strong scalability and robustness for complex, multi-hop reasoning tasks.

\paragraph{Impact of Retrieved Contexts (AdaComp-RAG):}

We vary the number of retrieved contexts $k \in \{1, 3, 5, 7, 10\}$ and report results in Table~\ref{tab:scalability_rag}. Accuracy peaks at $k=5$ (89.2\% EMA), balancing informativeness and efficiency. Higher values of $k$ introduce redundant or noisy content, increasing latency and memory usage with diminishing accuracy returns. The \texttt{Redundancy Detected} column estimates overlap via semantic similarity metrics, showing inefficiencies beyond $k=5$. These results validate the need for adaptive, selective retrieval to optimize reasoning quality and resource consumption.

\paragraph{Computational Analysis:}
We evaluate \textbf{HOLA} on GSM8K across Jetson Nano, Raspberry Pi 4, Intel i7-10700K, and NVIDIA A100 to assess deployment across diverse hardware. As shown in Table~\ref{tab:comp_analysis}, \textbf{HOLA} maintains high accuracy—87.6\% (Jetson Nano), 85.1\% (Raspberry Pi 4), and 89.2\% (A100)—demonstrating robust low-bit inference. Latency scales with compute: A100 is fastest (68 ms/sample), followed by Jetson Nano (841 ms) and Raspberry Pi 4 (1092 ms). Memory usage remains edge-compatible ($\sim$1.2 GB), and power stays low (7.5 W, 6.1 W) on edge devices. A100 delivers $14+$ samples/sec (97.2 W), suiting high-throughput workloads. \textbf{HOLA} proves deployable across edge and cloud, balancing speed, accuracy, and power. \textbf{\texttt{We focus on GSM8K due to its diverse, multi-hop numerical reasoning demands, offering a better testbed for latency and precision trade-offs than ARC.}}

\begin{figure*}[!t]
\vspace{-0.3cm}
    \centering
    \begin{subfigure}[b]{0.32\textwidth}
        \centering
        \includegraphics[width=\textwidth]{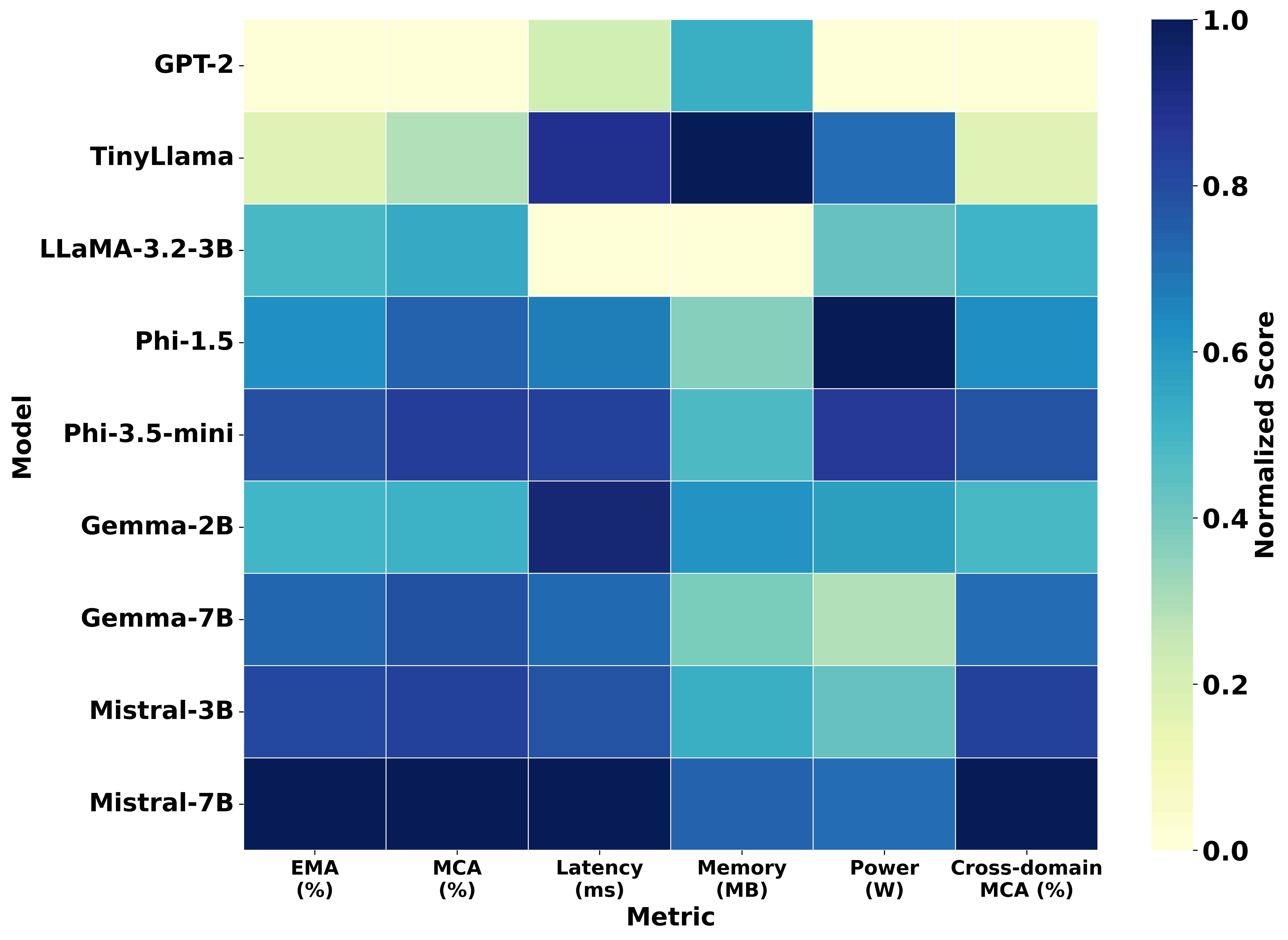}
        \caption{HOLA effect on model ranking}
    \end{subfigure}
    \hfill
    \begin{subfigure}[b]{0.32\textwidth}
        \centering
        \includegraphics[width=\textwidth]{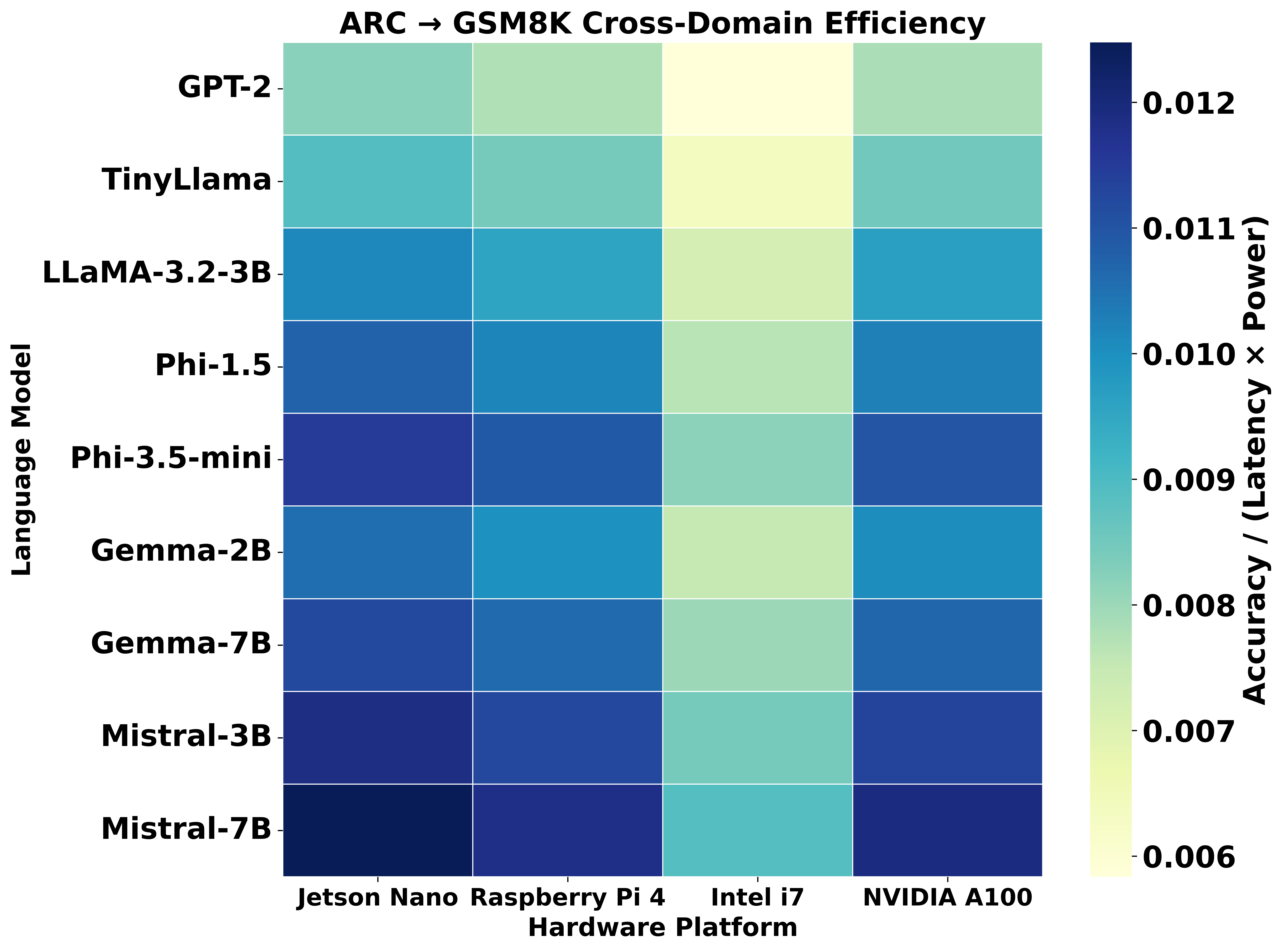}
        \caption{Transfer: ARC to GSM8K efficiency}
    \end{subfigure}
    \hfill
    \begin{subfigure}[b]{0.32\textwidth}
        \centering
        \includegraphics[width=\textwidth]{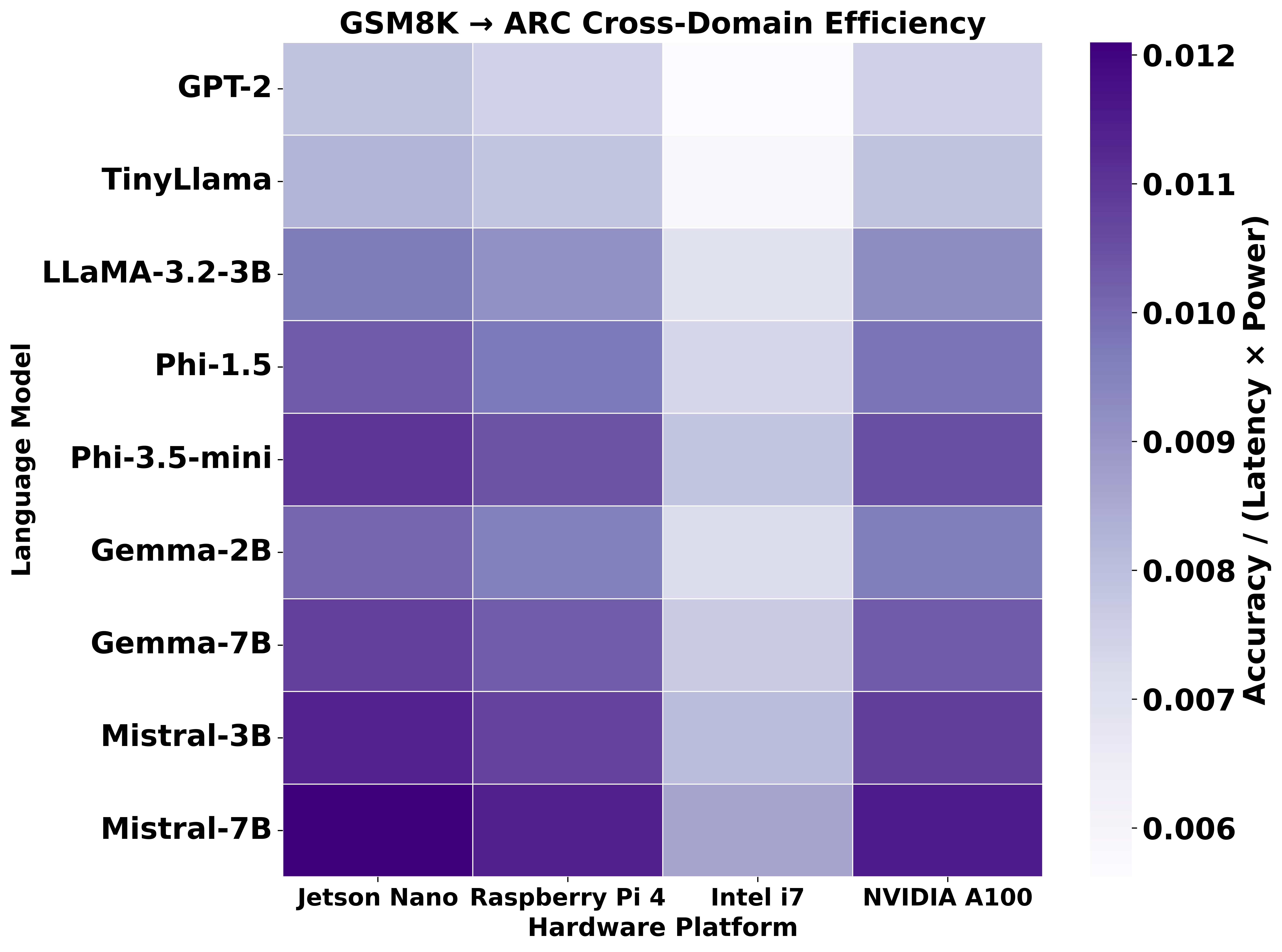}
        \caption{Transfer: GSM8K to ARC efficiency}
    \end{subfigure}
    \begin{subfigure}[b]{0.32\textwidth}
        \centering
        \includegraphics[width=\textwidth]{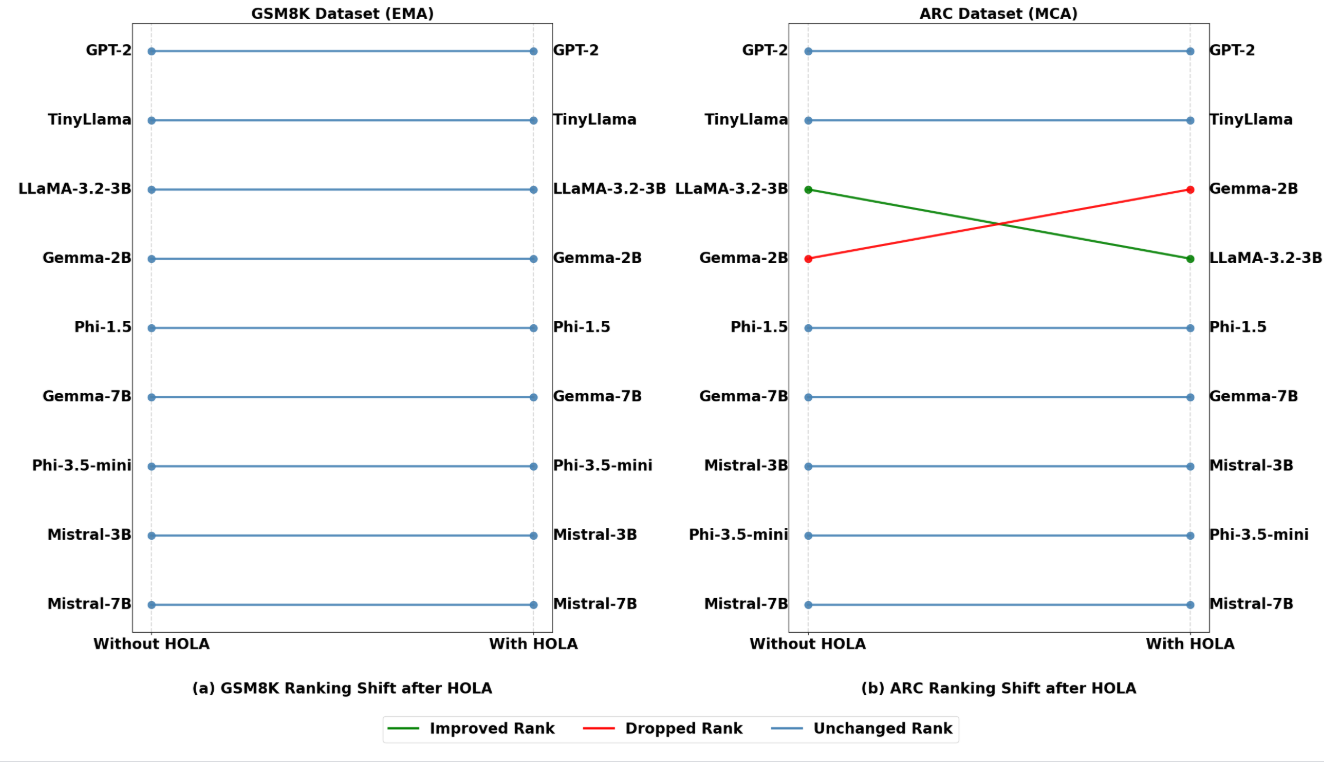}
        \caption{Ranking shifts due to HOLA}
    \end{subfigure}
    \hfill
    \begin{subfigure}[b]{0.32\textwidth}
        \centering
        \includegraphics[width=\textwidth]{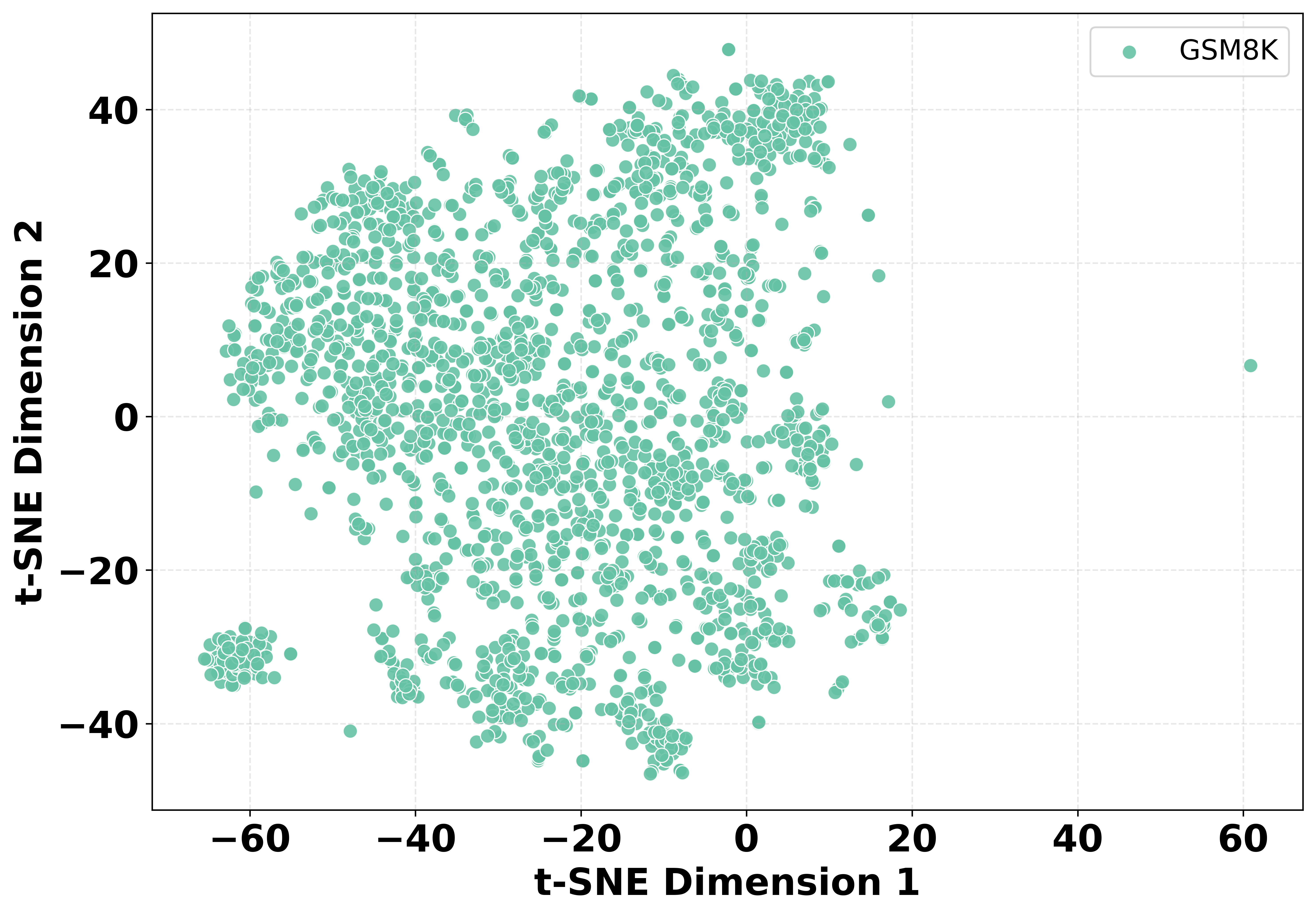}
        \caption{GSM8K latent space}
    \end{subfigure}
    \hfill
    \begin{subfigure}[b]{0.32\textwidth}
        \centering
        \includegraphics[width=\textwidth]{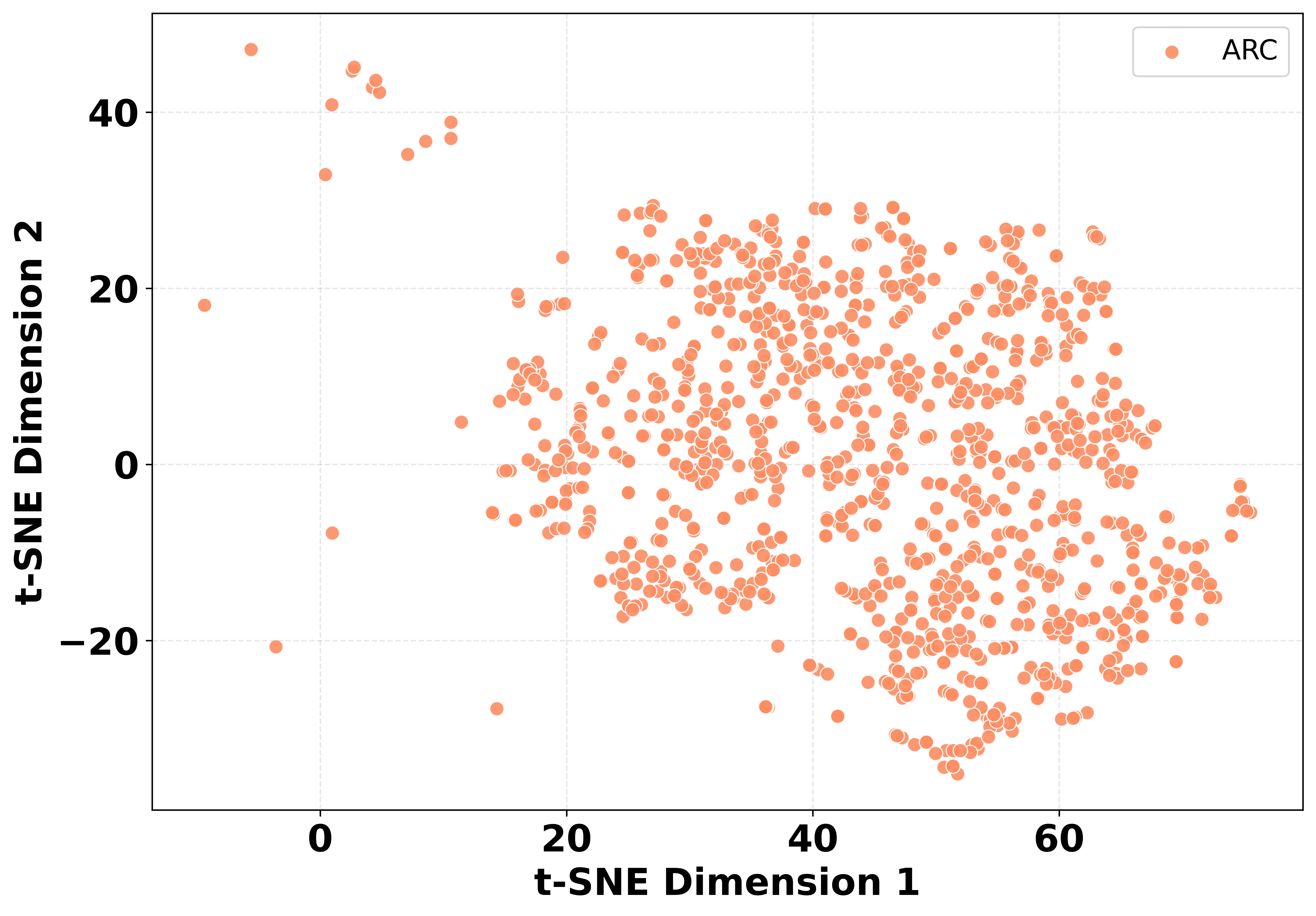}
        \caption{ARC latent space}
    \end{subfigure}
    \caption{Detailed analysis of the impact of HOLA optimization and domain transfer on large language models (LLMs) across tasks and hardware settings. (a) shows how HOLA influences model ranking for the GSM8K and ARC datasets. (b) and (c) illustrate the efficiency changes when transferring models between ARC and GSM8K tasks on various hardware. (d) compares ranking shifts caused by HOLA optimization. (e) and (f) provide t-SNE visualizations of the latent space to highlight task domain effects for GSM8K and ARC, respectively.}
    \label{Figure 2}
\end{figure*}

\paragraph{Qualitative Analysis} Figure \ref{Figure 2} provides a comprehensive qualitative analysis of the impact of \textbf{HOLA} optimization and domain transfer on LLMs across two benchmark tasks: ARC and GSM8K. Subfigure (a) illustrates how \textbf{HOLA} affects model rankings, showing that the optimization has a varied impact across different models and tasks—some models improve significantly, while others decline or remain unaffected. Subfigures (b) and (c) show the efficiency of cross-domain transfer from ARC to GSM8K and vice versa, respectively. The heatmaps reveal that transfer efficiency is highly dependent on model architecture and hardware configurations, with some models exhibiting strong generalization across tasks and others struggling. Subfigure (d) further highlights the shifts in model rankings caused by \textbf{HOLA}, indicating that optimization can reconfigure relative model performance on these benchmarks. Subfigures (e) and (f) use t-SNE to visualize the latent task spaces of GSM8K and ARC, respectively. The distinct clustering in each plot suggests that the two tasks occupy separate regions in the latent space, which helps explain the varying success of cross-task transfer. Together, these insights emphasize that both optimization and evaluation strategies must consider task specificity and domain effects to ensure robust performance across reasoning benchmarks.

\section{Conclusion and Future Work}
\label{sec:conclusion}

We propose \textbf{HOLA}, a hybrid framework for efficient retrieval-augmented generation, achieving strong accuracy-latency-memory trade-offs across edge and cloud platforms. Experiments validate its scalability and deployment viability. Future work includes multilingual retrieval, long-context reasoning, hardware-aware training for ARM/RISC-V, and real-time integration in low-power systems like dialogue agents and embedded QA bots.

\section*{Limitations}
\label{sec:Limitations}

While \textbf{HOLA} delivers strong performance-efficiency trade-offs, several limitations remain. The retrieval pipeline depends on static indexing, limiting adaptability to dynamic knowledge or temporal queries. Aggressive compression, while beneficial for edge deployment, can introduce non-trivial accuracy loss on ultra-low-memory devices. Current evaluation is constrained to GSM8K, and generalizability to more complex, multi-turn, or multi-modal tasks remains unverified. Robustness under adversarial or noisy retrievals is also underexplored. Additionally, real-world deployment factors—such as thermal throttling, energy spikes, or long-term device wear—are not modeled. 

\section*{Ethics Statement}
\label{sec:Ethics Statement}

This work follows established ethical guidelines in AI development and deployment. All datasets used are publicly available, de-identified, and compliant with data usage norms. \textbf{HOLA} was designed with efficiency and accessibility as core principles, targeting responsible deployment in resource-constrained and educational contexts. While the model improves reasoning automation, we recognize potential risks such as misuse for generating inaccurate content. Additionally, fairness and transparency remain active concerns. Responsible release protocols, including usage documentation and bias monitoring, will guide the deployment of \textbf{HOLA} in real-world scenarios.

\bibliography{anthology,custom}
\bibliographystyle{acl_natbib}

\end{document}